# COMPUTER VISION BY YOLOs FOR DETECTION OF MAJOR DISEASES IN VIETNAMESE CUSTARD APPLE USING FRUIT IMAGERY

**Tri Nhut Do[1,2,*],Quoc Tuan Vo[3]**

*[1]University of Information Technology (UIT)*

*[2]Vietnam National University in HoChiMinh City (VNUHCM), Ho Chi Minh City, Vietnam*

*[3]Faculty of Information Technology, University of Phan Thiet,*

*Phu Thuy Ward, Lam Dong Province, Vietnam*

*Email: *trinhutdo@uit.edu.vn*

## ABSTRACT

Amid Vietnam's rapid digital transformation in agriculture, the integration of artificial intelligence (AI) technologies for identifying, monitoring, and predicting crop pests and diseases has emerged as an inevitable trend. Custard apple (*Annona squamosa* L.), a high-value fruit crop, suffers significant losses due to prevalent diseases, including Anthracnose, Black Canker, Diplodia rot, Leaf Spot, and Mealybug infestations. This paper proposes a robust computer vision framework that leverages state-of-the-art YOLO (You Only Look Once)-series object detection models to enable early and accurate detection of major diseases on custard apple leaves and fruits. The proposed system is designed for practical deployment through mobile and edge-device applications in the future, empowering farmers with timely, actionable diagnostics to minimize yield losses. Furthermore, we systematically evaluate and benchmark the performance of the latest YOLO variants (YOLOv8, YOLOv11, and YOLOv12) on a large-scale, real-world custard apple disease dataset collected under diverse field conditions in Vietnam. This work not only delivers a high-performance, deployable solution but also lays the groundwork for establishing a comprehensive digital database to advance AI-driven agricultural research in Vietnam.

## 1. INTRODUCTION

In the context of the strong digital transformation of agriculture in Vietnam, the application of artificial intelligence in the identification, monitoring, and forecasting of plant pests and diseases has become an inevitable trend. Many fruit trees with high economic value, such as custard apple, are being severely affected by diseases such as anthracnose, black rot, Diplodia mold, leaf spot, and mealybug. These diseases are often difficult to detect early with the naked eye, especially in large-scale production conditions, leading to reduced productivity and fruit quality [1]. Traditional methods of diagnosing plant diseases, based on manual observation or farmers' experience, do not ensure consistency and high accuracy. Meanwhile, the development of Deep Learning models, especially Convolutional Neural Networks (CNN), has opened up a new approach to automatically detecting and classifying plant diseases through images [2]. Among modern deep learning methods, the YOLO (You Only Look Once) model is one of the most prominent architectures in the field of Object Detection because of its fast, accurate, and compact detection capabilities. New versions such as YOLOv11 and YOLOv12 are continuously improved in terms of network architecture, Attention mechanism, and inference speed, making the model capable of being deployed on resource-limited devices such as phones or agricultural drones [3].

Based on the urgent practical needs mentioned above, we conducted research and proposed a system that applies AI technology with YOLOs algorithm models to detect some common diseases on custard apples. As a result, the proposed system is implemented and many experiments are conducted and their results collected in order to:

- Evaluate and compare the performance of new generation YOLO models in detecting plant diseases. The analysis aims to determine the model with the best performance, suitable for practical application.
- Propose an AI-driven practical application pipeline to help farmers detect early and prevent plant diseases effectively, aiming at deployment on mobile devices or farmer support systems.
- Contribute to building a digital database to serve agricultural AI research in Vietnam.

In this paper, the research focuses on detecting and classifying common diseases on custard apple (*Annona squamosa* L.) through real-life photos taken in natural conditions in Vietnam. The disease groups are selected based on the frequency of occurrence and economic impact, including:

- Anthracnose
- Black Canker
- Diplodia Rot
- Leaf Spot
- Mealy Bug

Each disease group has a corresponding image dataset. All images are manually labeled and saved in a YOLO label format, to serve for the training and testing in further [4].

Therefore, the system proposed in this paper not only has scientific significance in applying deep learning technology, but also has practical significance in improving the efficiency of crop management, towards smart and sustainable agriculture [5].

The remainder of this paper is organized as follows: Section 2 provides an overview of current research on AI-driven agriculture, Section 3 introduces the proposed system design and its implementation, Section 4 presents quantitative assessment results and evaluation, and Section 5 concludes the paper with future directions.

## 2. CURRENT RESEARCH ON AI-DRIVEN IN AGRICULTURE

### 2.1. Overview

Over the past decade, artificial intelligence (AI) has become the core technology platform of the 4.0 industrial revolution, with the ability to simulate human intelligence through machine learning and deep learning algorithms [6]. In the agricultural sector, AI does not only stop at automating production processes, but also supports image analysis, weather forecasting, plant disease diagnosis, and yield monitoring [7]. This is depicted in Fig. 1.

The combination of AI and computer vision has especially opened up many new applications, such as plant variety classification, pest detection, growth monitoring, and fruit quality inspection. Remarkable advances in deep learning - especially the advent of convolutional neural networks (CNNs) - have given computers the ability to "see" and distinguish image features with higher accuracy than humans in many cases.

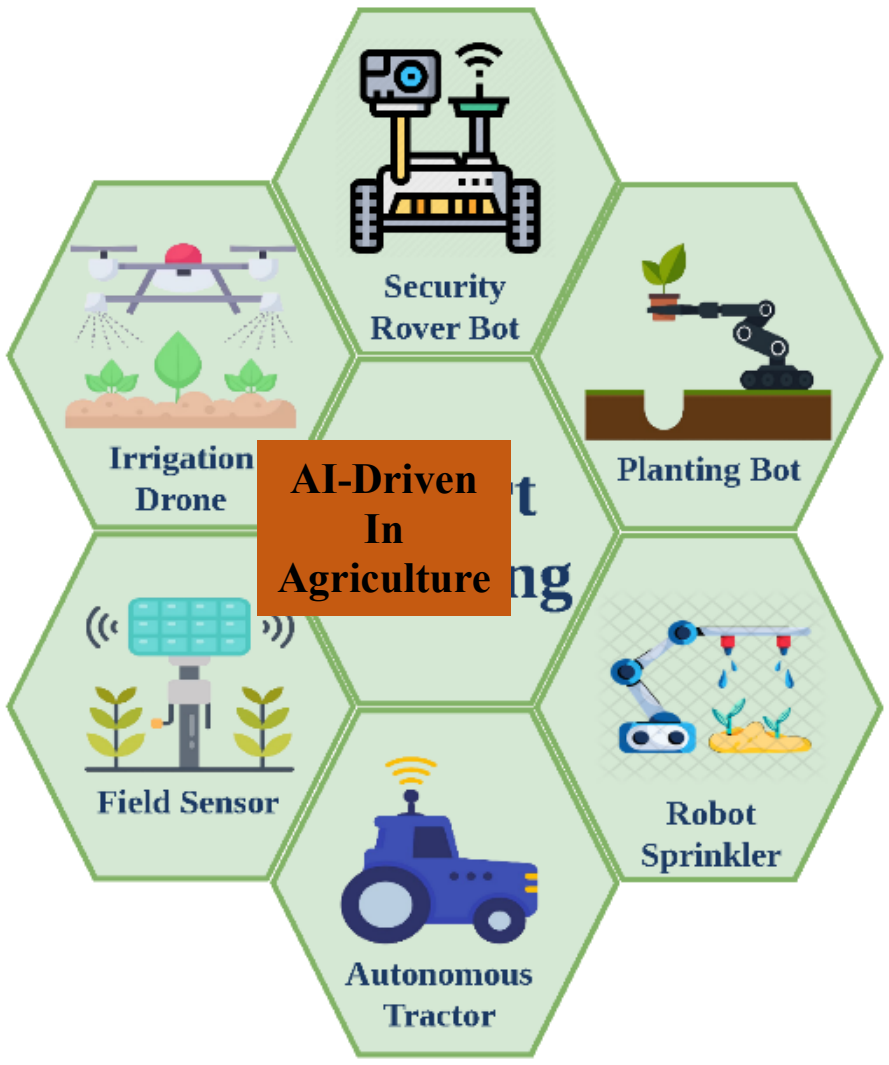

*Fig. 1.* Current research on AI-driven in agriculture.

In Vietnam, the application of deep learning models in agriculture is still in its infancy, but there have been many potential studies such as: disease detection on rice, coffee, banana, and recently custard apple - a crop with high economic value in the southern provinces [8]. However, most of the current research only stops at the level of disease classification through static images. While the problem of detecting objects includes both determining the location of the disease and classifying the disease area, it requires more powerful models. Typical such models include the YOLO (You Only Look Once) family model, the SSD (Single Shot Multibox Detector) model, and the Faster R-CNN model. In particular, YOLO models are considered pioneering models for detecting plant diseases in the natural environment because of their outstanding advantages, such as fast inference ability, compact structure, and direct deployment on mobile devices. Comparing the performances when applying the latest YOLO versions, including YOLOv8, YOLOv11, and YOLOv12, in the context of Vietnamese agricultural data, therefore, has special significance both scientifically and practically [9].

### 2.2. Object Detection Model Classification

Object Detection Algorithms Based on Artificial Intelligence, or known as AI-driven object detection, is one of the core problems of Computer Vision. Unlike the Image Classification Algorithms that only label the entire image, the Object Detection Algorithms require the model to determine both the location (bounding box coordinates) and the type of object (class) in the same image [10]. Modern deep learning models use convolutional neural networks (CNN) to extract image features, then combine inference layers to predict the object's location. This problem is widely applied in security, transportation, healthcare, and especially smart agriculture, where it is necessary to detect pests, count plants, or determine crop status.

AI-driven object detection models are divided into two main groups: two-stage detection models and one-stage detection models [11]. This is depicted in Fig. 2.

*1) The two-stage detection model group is traditional detectors, consisting of two steps:*

- *Region Proposal*, where the model finds regions that are likely to contain objects) and,
- *Classification and location refinement*, where accurately determine labels and bounding boxes).

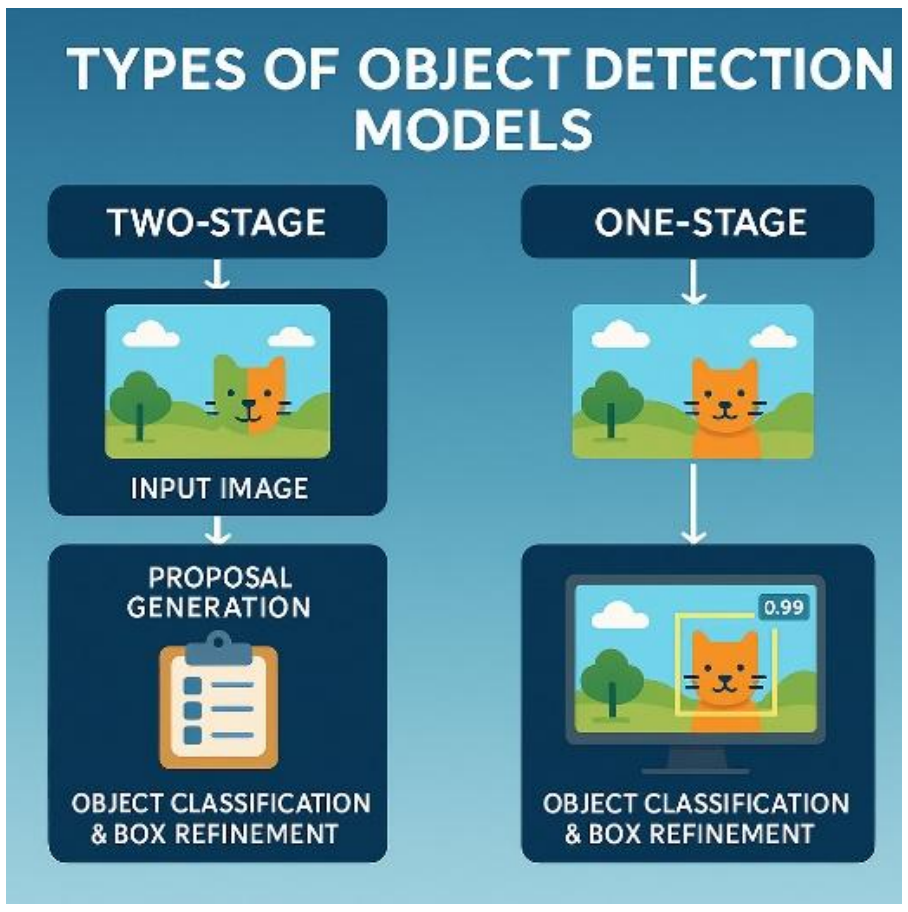

*Fig. 2.* Object detection models.

- *Typical representatives:* R-CNN (2014) [12], Fast R-CNN and Faster R-CNN (2015) [13]. Among them, the Fast and Faster ones improved speed by integrating the Region Proposal Network (RPN) layer, enabling end-to-end learning.
- Advantages: High accuracy, effective with small objects.
- Disadvantages: Slow inference speed, not suitable for real-time requirements.

2) *The one-stage detection model group:*

- Detectors in this group analyze the input image in a single pass, bypassing the Region Proposal step to directly detect objects.
- Prominent examples of such models include You Only Look Once (YOLO), Single Shot MultiBox Detector (SSD), and RetinaNet.
- Advantages: Fast, suitable for real-time applications.
- Disadvantages: Lower accuracy than the two-stage group when detecting very small or obscured objects.

The primary difference between these two methods lies in their use of RPN; specifically, two-stage detectors generate region proposals, whereas one-stage detectors do not. Although two-stage detectors are generally more accurate due to their architecture, recent advancements in one-stage detectors have substantially narrowed the performance gap (as of 2025, with the release of YOLO v12). Ultimately, the decision between one-stage and two-stage detectors should be guided by specific application requirements rather than a straightforward performance trade-off.

### 2.3. AI-Driven Research on Applications in Agriculture

Studies mainly used traditional CNN networks such as AlexNet, VGG16, and ResNet to classify disease images. Among them, Mohanty et al. (2016) [14] is one of the pioneering works, with a CNN model trained on more than 50,000 plant disease images in the PlantVillage dataset, achieving an accuracy of over 99%. However, the limitation of this research is that it cannot determine the location of the diseased area, only labeling the whole image.

Research on object detection models such as YOLOv3, YOLOv5, SSD, and EfficientDet has created a new turning point. Eman et al. (2024) [15] applied YOLOv4 to detect plant leaf disease and achieved mAP@0.5 = 0.89, outperforming Faster R-CNN in both speed and accuracy. Li et al. (2023) [16] improved YOLOv5 using an attention mechanism (SE module) to increase 6% accuracy on the grape leaf disease dataset. Liu et al. (2023) [17] integrated Focal Loss and Mosaic augmentation into YOLOv5 to detect tomato disease, achieving mAP50 = 0.93.

New YOLO versions such as YOLOv8, YOLOv11, and YOLOv12 were developed by Ultralytics in the direction of Anchor-Free, ConvNeXt backbone, and Global Context Attention, which improves the ability to detect small and complex disease areas. Rahima Khanam and Muhammad Hussain (2025) [18] showed that YOLOv12s outperformed YOLOv8s in mAP (0.48 vs. 0.44) in detecting leaf disease on lemon trees, while reducing the inference time by 30%. These results demonstrate that the new YOLO models are suitable for real-life agricultural environments where lighting conditions and image backgrounds change continuously.

In Vietnam, research on AI-driven applications in agriculture has been growing strongly in the past 5 years, focusing on plant disease diagnosis and smart crop management. Trinh et. al. (2023) [19] applied the YOLOv5s model to identify diseases on rice leaves, achieving 93% accuracy, proving the feasibility of YOLO in Vietnamese data. Moreover, some research groups at the Vietnam Academy of Agriculture and Can Tho University have also tested combining AI with UAVs (drones) to collect plant disease images, opening up an "AI + IoT" approach in digital agriculture.

### 2.4. Evaluation

Synthesizing the mentioned studies/researches/works, we obtain some main conclusions:

- YOLO models are still the most optimal solution for the problem of detecting plant diseases in natural conditions.
- Previous studies focused on rice, tomato, grape, orange; while *Annona squamosa* (Custard Apple) has not been exploited deeply.
- The research gap lies in standardizing the data - model - evaluation pipeline specifically for Vietnamese agriculture.
- This topic inherits previous approaches, but focuses specifically on *Annona squamosa*, and directly compares 3 generations of YOLO to evaluate the practical applicability.

## 3. SYSTEM DESIGN

Based on the mentioned evaluation, a detection system is designed according to the applied - experimental - optimization approach, specifically as follows:

- Building a field data set of diseases in Binh Duong.
- Collecting images directly from gardens, including many lighting conditions and complex backgrounds.
- Manually labeling with *LabelImg* and standardizing the format according to YOLO (class, x_center, y_center, width, height).
- Preprocessing and data augmentation (Data Augmentation).

Implementing augmentation techniques such as flipping, rotating, adding noise, and converting to grayscale (grayscale 10%), applied on the *Roboflow* platform to balance the data of disease classes.

- Training three models, YOLOv8s, YOLOv11s, YOLOv12s, on the same data set with the following configuration:

Training configuration: epoch = 100, batch = 4, imgsz = 960, optimizer = SGD.

Using GPU T4 (Google Colab) to speed up the training process.

Record all log, loss, precision, recall, mAP50, mAP50–95.

- Compare performance and evaluate quantitatively using the following indicators:

$$Precision = TP / (TP + FP)$$

$$Recall = TP / (TP + FN)$$

$$F1 - score = 2 \times (Precision \times Recall) / (Precision + Recall)$$

$$mAP@0.5 = (\Sigma\ AP_i) / N$$

where:

TP: True Positive

FP: False Positive

FN: False Negative

AP_i: Average Precision of the $i^{th}$ class

N: number of classes.

### 3.1. Data Collection and Construction

- Data source: photos of custard apples in gardens in Binh Duong (period 2024–2025), including disease conditions: Anthracnose, Black Canker, Diplodia Rot, Leaf Spot, Mealy Bug. The photos were taken with digital cameras and smartphones, under different lighting, humidity, and background conditions to accurately reflect the natural environment. A photo example data source is illustrated in Fig. 3.

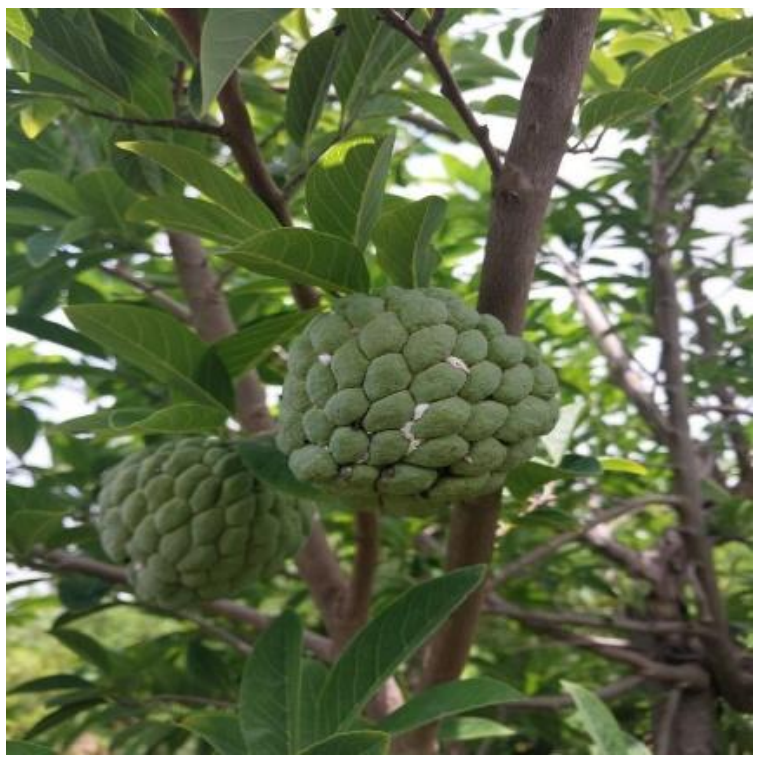

*Fig. 3.* Photo data source.

- The dataset includes 5 common disease classes that commonly appear on custard apple. They are listed in Table 1.

*Table 1.* Disease classes

| Disease Name | Label Number | Short Description |
|---|---|---|
| Anthracnose | 0 | Round dark brown dents, usually on the fruit peel or leaf. |
| Black Canker | 1 | The damaged area is dark, spreading around the fruit stem. |
| Diplodia Rot | 2 | Causes rot in the peel and flesh of the fruit, often accompanied by white mold. |
| Leaf Spot | 3 | Round yellow-brown spots, concentrated on the leaf surface. |
| Mealy Bug | 4 | Appear in clusters of white cotton, clinging to the stem and leaf stalk. |

- Standard size: 960 × 960 pixels. Each image is standardized to 960×960 pixels and saved in JPG or PNG format.
- Number of samples: 996 original photos, including those enhanced by data augmentation techniques [20].
- Objective: ensure diversity in shooting angles, lighting and colors, helping the model learn disease characteristics in natural conditions.
- Data size and split ratio are listed in Table 2.

*Table 2.* Data size and split ratio

| Data for | Data Size | Ratio | Usage Note |
|---|---|---|---|
| Train | 796 photos | 80% | Used to train the model |
| Validation | 200 photos | 20% | Used to validate and calculate mAP |
| Test | -- | -- | Test the model on images outside the dataset |

### 3.2. Data Preprocessing and Labeling

- Image preprocessing:
  - Convert to RGB format.
  - Normalize size and aspect ratio.
  - Filter out blurry, underexposed, or mislabeled images.
- Object labeling:
  - Use *LabelImg* and *Roboflow* to draw bounding boxes.
  - Save labels in YOLO format: *class x_center y_center width height*
  - Create *data.yaml* configuration file containing the number of classes and the train/val path.
- Data augmentation:
  - Rotate images ±20°, crop, flip.
  - Add Gaussian noise, adjust brightness.
  - Grayscale 10% and add random noise to 0.5% of pixels.

### 3.3. Model Training

- Process: Train YOLO models v8s, v11s and v12s in turn.
- Objective: compare 3 modern YOLO models - YOLOv8s, YOLOv11s, and YOLOv12s.
- Training tool: *Ultralytics* YOLO library (v8.3.220) on *Google Colab* GPU T4 (16GB) environment.
- Main training parameters are listed in Table 2.

*Table 3.* Training parameters

| Parameter | Value | *Parameter* | Value |
|---|---|---|---|
| Epochs | 100 | Optimizer | SGD |
| Batch size | 4 | Confidence threshold | 0.25 |
| Image size | 960 | IOU threshold | 0.5 |

- Average training time is 2 hours for each version of YOLOs.

### 3.4. Performance Evaluation and Comparison

The results are presented through:

- Precision, Recall, F1, mAP@0.5, mAP@0.5–0.95 charts
- Loss (train/val) charts
- Confusion Matrix
- PR Curve and F1–Confidence Curve

## 4. QUANTITATIVE ASSESSMENT RESULTS AND EVALUATION

### 4.1. Performance Comparison

Three models, YOLOv8s, YOLOv11s, and YOLOv12s, are trained in parallel on the same dataset of 996 images of 5 common diseases on custard apple.

Overall assessment of the Training Results shows that:

- All three models converge stably after about 80 epochs.
- YOLOv12s achieves the highest loss stability, with no signs of overfitting.
- Table 4 is the quantitative assessment results.

*Table 4.* Quantitative assessment results

| YOLOs | mAP@0.5 | Precision | Recall |
|---|---|---|---|
| YOLOv8s | 0.842 | 0.878 | 0.818 |
| YOLOv11s | 0.862 | 0.890 | 0.861 |
| YOLOv12s | 0.904 | 0.895 | 0.888 |

The results show that YOLOv12s achieves the highest mAP@0.5 (0.904) and the best Precision (0.895), confirming the superior performance of backbone ConvNeXt + Global Context Attention in extracting small disease region features.

*Evaluation*

- YOLOv12s has the highest value in all three indicators.
- YOLOv8s has relatively high Precision, but low Recall → easily misses diseased areas.
- YOLOv11s is intermediate, but does not achieve as high mAP as YOLOv12s.
- In addition to the two presented indicators in Table 4, other criteria are also quantified, and the final results are evaluated as listed in Table 5.

Analysis of Table 5 concludes as follows:

- The YOLOv12s model provides the best results in the problem of disease detection on custard apple, balancing accuracy, generalization ability, and speed.
- Although the hardware requirements are higher than YOLOv8s, the superior performance makes YOLOv12s the most suitable candidate for practical applications in smart agriculture.

*Table 5.* YOLOv12s indicators

| Criteria | YOLOv8s | YOLOv11s | YOLOv12s |
|---|---|---|---|
| Overall mAP | Average | Fair | Best |
| Precision | High | Average | Highest |
| Recall | Low | Average | Fairly Good |
| Generalization | Average | Fairly | Stable |
| Inference Speed | Fast | Average | Slightly Slower |
| Real-world Deployability | Good | Good | Very Good |

## 4.2. YOLOv12s results

Therefore, the YOLOv12s model was selected for experimental development. The indicators of YOLOv12s showing the accuracy of each disease class are listed in Table 6.

*Table 6.* YOLOV12s indicators

| Disease Class | Precision | Recall |
|---|---|---|
| Anthracnose | 0.85 | 0.90 |
| Black Canker | 0.72 | 0.68 |
| Diplodia Rot | 0.70 | 0.74 |
| Leaf Spot | 0.81 | 0.79 |
| Mealy Bug | 0.88 | 0.85 |

*Evaluation*

- The Anthracnose and Mealy Bug classes had the highest accuracy.
- The Black Canker and Diplodia Rot classes were slightly confused due to the similar color of the lesions.
- In general, the model achieved an average class accuracy of > 80%.

Some results of disease detection on custard apple using YOLOv12s are shown in Fig. 4.

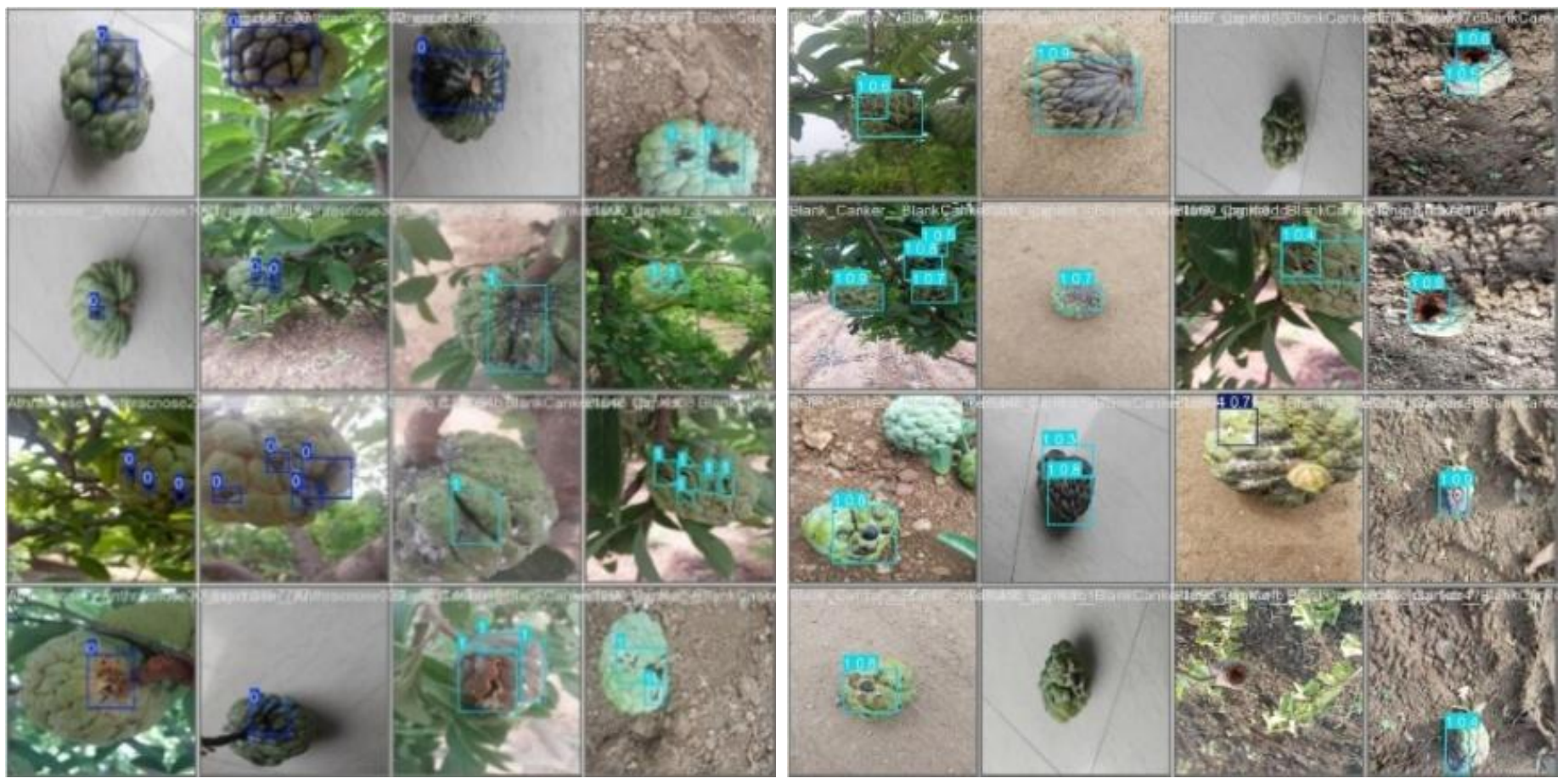

*Fig. 4.* Disease detection results along with numbers representative to specify diseases listed in Table 1

## 5. CONCLUSION

In this paper, a disease detection system for custard apple is designed based on artificial intelligence. The design is meticulously carried out in each step on the same data set applied to the 3 best versions of YOLO, v8s, v11s, and v12s, respectively. The final training results demonstrate that the YOLOv12s version is the most optimal. From there, experiments on real images in addition to the training image set have produced test results confirming that: diseased areas are accurately delineated by color frames (bounding box), disease labels and confidence rates are clearly displayed, YOLOv12s does not detect false positives in the background. Experimental results also show that on average, each image detects 2–3 diseased regions correctly, with inference accuracy reaching nearly 88% on images outside the training set and an average prediction time of 18.3 milliseconds per image with a GPU T4 of Google Colab. Once again, the experimental results demonstrate that YOLOv12s confirms the effectiveness of ConvNeXt Backbone and Global Context Attention in the complex disease region detection problem.

With the high accuracy achieved, in the future, the YOLOv12s model can be deployed on mobile applications, agricultural surveillance cameras, or AI drones to help farmers detect diseases early, reduce yield loss and medicine costs, and contribute to realizing the goal of digital agriculture and digital transformation in the field of cultivation in Vietnam.

However, our research team needs to expand the dataset further than the 996 images in this study (e.g., over 5000 images); at the same time, it is necessary to combine Transfer Learning and Fine-tuning Attention Map techniques to reduce training time; and deploy the YOLOv12s model on Edge AI devices (NVIDIA Jetson / Coral TPU) for practical use.

**Acknowledgment:** This research is funded by the University of Information Technology-Vietnam National University HoChiMinh City. This research couldn't have been done without the help of graduate student Nguyen Tan Dat’s support.